\documentclass{article}
\usepackage{spconf,amsmath,graphicx}

\usepackage{graphicx}
\usepackage{nicefrac}       
\usepackage{microtype}      
\usepackage{lipsum}
\usepackage{amsmath}
\usepackage{graphicx}
\usepackage{float}
\usepackage{amsthm}

\usepackage{bm}
\usepackage{amsfonts}
\DeclareMathOperator*{\argmax}{arg\,max}
\DeclareMathOperator*{\argmin}{arg\,min}
\usepackage{multirow}
\usepackage{enumerate}

\title{Focused Discriminative Training For Streaming  CTC-Trained  Automatic Speech Recognition Models}
\name{Adnan Haider*, \thanks{*Corresponding author.} Xingyu Na, Erik McDermott, Tim Ng, Zhen Huang and Xiaodan Zhuang.  }
\address{Apple Inc., \\
$\{$adnan\_haider, na\_xingyu, erik\_mcdermott, tim\_ng, zhen\_huang, xiaodan\_zhuang$\}$@apple.com.}

\begin{document}
%
\maketitle

\begin{abstract}
This paper introduces a novel training framework called Focused Discriminative Training (FDT) to further improve streaming word-piece  end-to-end (E2E) automatic speech recognition (ASR) models  trained using either CTC or an interpolation of CTC and attention-based encoder-decoder (AED) loss. The proposed approach presents a novel framework to identify and improve a model's recognition on challenging segments of an audio. Notably, this training framework is independent of hidden Markov models (HMMs) and lattices, eliminating the need for substantial decision-making regarding HMM topology, lexicon, and graph generation, as typically required in standard discriminative training approaches.  Compared to additional fine-tuning with MMI or MWER loss on the encoder, FDT is shown to be more effective in achieving greater reductions in Word Error Rate (WER) on  streaming models trained on LibriSpeech. Additionally, this method is shown to be effective in further improving a converged word-piece streaming E2E model trained on 600k hours of assistant and dictation dataset. 
\end{abstract}

\section{Introduction}

The combination of connectionist temporal classification (CTC) \cite{ctc} and attention-based encoder-decoder (AED)\cite{aed1,aed3,aed2} loss has become a powerful approach for training end-to-end (E2E) models to address both streaming and non-streaming speech recognition tasks \cite{orginalCTCAED, wenet,wenet2}.


This paper in particular aims to address the challenging task of further improving a streaming word-piece ASR model that has converged with respect to CTC loss. For neural network models whose output units correspond to phones, sequence discriminative training approaches such as MMI, sMBR, EMBR, etc. \cite{MMI,kingsbury2012, MattShannon}, which embed the models within an HMM \cite{Baum1967} framework, have been shown to be effective in achieving further WER improvements. However, the implementation of discriminative sequence training is a time-consuming and challenging process. Implementations of both lattice-based \cite{MMI,MPE} and lattice-free sMBR/MMI  \cite{LFMMI,mitchel,tian2022integrating} techniques necessitate careful design decisions regarding the choice of HMM topology, lexicon, and graph generation \cite{wfst}. This results in a slower evolution of the training pipeline compared to the rapidly evolving training frameworks for E2E models. In particular, lattice-based MMI introduces an additional computational overhead by requiring pre-generated lattices, potentially diminishing the training framework's flexibility.  Furthermore, in the LF-MMI paper \cite{LFMMI}, the authors acknowledge the difficulty of further improving models trained with CTC and thus employ their discriminative training approach to fine-tune models initially trained with Cross-Entropy loss. The current variant of LF-MMI trains models with phonetic output units from scratch. It should be noted that, aside from the mentioned approaches, the Minimum Word Error Rate (MWER) \cite{MWER} loss has been shown to be effective when applied on top of the decoder in encoder-decoder based models. This method is an HMM-independent approach.  However, at the time of writing this paper, this work found no evidence in the literature that presents a successful application of this loss to fine-tune word-piece based CTC-trained encoders using N-best lists \cite{MBRT}.

This paper introduces a novel approach called Focused Discriminative Training (FDT) to specifically fine-tune  encoders with word-piece outputs trained using either CTC or an interpolation of CTC and AED (CTC+AED) loss. The motivation behind the algorithm is to address a particular problem encountered by streaming E2E ASR models. To exemplify the problem, consider the following voice assistant request: "Hey Assistant, call Beethoven." Assume that in the dataset used to train the model, there are ample audio examples for the phrase "Hey Assistant, call," but few examples for the proper noun "Beethoven." In such situations, the intended action is correctly recognised by the ASR model, but difficulties are encountered in identifying the intended callee. Current discriminative training approaches use sequence-level objectives, which aim to increase the probability of the entire word sequence as a whole without specifically targeting the regions where ASR mis-recognises words. Ideally, a framework that fine-tunes the model specifically on these challenging regions is desired.

The  FDT algorithm consists of 3 key components: implicit data selection, an error region detection mechanism and a novel constrained CTC-based contrastive segment-level loss. FDT integrates these components to identify and improve the recognition of challenging segments within the input sequence. The proposed method is HMM-independent, eliminating the requirement for various components, like HMM topology, lexicon, and utilises N-best lists. Thus, the method is also lattice-free. 

The efficacy of the algorithm is demonstrated through further fine-tuning  streaming encoders with word-piece outputs, trained on the LibriSpeech dataset using CTC and CTC+AED loss. The method's efficacy was compared with MMI and MWER loss applied on top of the encoder. The latter two approaches can be viewed as the CTC versions of previous work that utilised N-best lists for MWER \cite{MWER,MBRT,MBRLM} and MMI training \cite{MWERAED,MBRLM} of word-piece based AED and Neural Transducer models. To the best of our knowledge, this work presents the first application of MWER on word-piece based CTC-trained models. The MMI and MWER experiments reported here do not use a separate language model.  This study also assesses the  FDT algorithm’s efficacy in improving a converged model trained on a large-scale dataset from a distinct domain. Results demonstrate that when applied to further fine-tune a converged word-piece streaming E2E model trained on a dataset comprising 600K hours of assistant and dictation data with CTC+AED loss, the algorithm achieves additional improvements in WER. The rest of this paper is organised as follows: Sec.4 describes the proposed method. Sec.5 details the data, model, and experimental results. Conclusions are provided in Sec.6.

\section{Notation}
To facilitate understanding of the mathematical statements in this work, let
\begin{enumerate}[i]
     \item $ P_{\bm{\theta}} (\cdot)  = P(\cdot| \bm{\theta})$ - Probability function dependent on the choice of model parameters.
     \item $\mathbf{x}_{1:T}$ denote an input sequence of feature vectors, $\mathbf{W}^r_{1:k}$ denote its associated reference word sequence with $\mathbf{l}^r_{1:U}$ representing corresponding target word-piece label sequence. The superscript $r$ denotes `reference'.
      \item $\bm{e}_t = f(\bm{x}_{1:t}, \bm{\theta})$ denote the mapping of an acoustic encoder that strictly operates on frames corresponding to the left context for a fixed choice of model parameters $\bm{\theta}$.
      \item   $P_{\bm{\theta}}( \bm{z}_{t} |\mathbf{x}_{1:t}) = { \tt softmax}(\bm{e}_t)$ denote the conditional distribution at time $t$  yielded by the acoustic encoder where $\bm{z}_{t}$  represents a token/blank realisation  at time $t$.
      \item   ${1}_{\mathbf{l}_{1:U}}(\bm{z}_{1:T})$ be the characteristic function that equals 1 if and only if $\mathbb{B}(\bm{z}_{1:T}) = \mathbf{l}_{1:U}$  where $\mathbb{B}(\cdot )$  presents  a deterministic mapping that removes all blanks and repetitive tokens but preserves the repetitive tokens separated by blanks. 
\end{enumerate}

\section{The CTC Model \label{sec:ctc_model}}

Given an input-word-piece label sequence pair $(\mathbf{x}_{1:T}, \mathbf{l}_{1:U})$, CTC training implicitly constructs a specific form of a discriminative latent variable probabilistic model for the conditional distribution $P({\mathbf{l}}_{1:U}| \mathbf{x}_{1:T})$ from the frame-level conditional distribution $P_{\bm{\theta}}( \bm{z}_{t} |\mathbf{x}_{1:t})$ as follows:

\begin{align}
   F_{\mathrm{CTC}}(\bm{\theta})  &=  -\log P_{\bm{\theta}}({\mathbf{l}}_{1:U}| \mathbf{x}_{1:T}),  \notag\\ 
   &= -\log  \  \sum_{\bm{z}_{1:T}}    P_{\bm{\theta}}(\bm{z}_{1:T}| \mathbf{x}_{1:T}) \  \mathbf{1}_{{\mathbf{l}}_{1:U}}(\bm{z}_{1:T} ),   \label{eqn:ctclat} \\
   &=  -\log  \  \sum_{\bm{z}_{1:T}}  \Big [ \prod_t^T  P_{\bm{\theta}}( \bm{z}_{t} |\mathbf{x}_{1:t}) \Big ]     \mathbf{1}_{{\mathbf{l}}_{1:U}}(\bm{z}_{1:T} )  \label{eqn:ctccond}
\end{align}
Eqn(\ref{eqn:ctclat}) illustrates how the restriction imposed by the characteristic function  gives rise to the implicit discrete latent variable  model, while eqn(\ref{eqn:ctccond}) corresponds to the CTC independence assumption. Conditioned  on this specific choice of discriminative model, acoustic encoder mapping $\bm{e}_t$, and model parameters $\bm{\theta}$, 
\begin{align}
\hat{\bm{z}}_{1:T} (\bm{\theta}) = \argmax_{\bm{z}_{1:T}: \mathbb{B}(\bm{z}_{1:T})    = \mathbf{l}_{1:U}} \prod_t^T  P_{\bm{\theta}}( \bm{z}_{t} |\mathbf{x}_{1:t})
\end{align}
corresponds to  the  CTC alignment  generated by the model for the input-label sequence pair $(\mathbf{x}_{1:T}, \mathbf{l}_{1:U})$.

\section{FDT algorithm \label{sec:FDT}}
 This work introduces a two-stage iterative algorithm called FDT-`Focused Discriminative Training'. The algorithm consists of 3 essential components: implicit data selection coupled,  an error region detection system and a CTC based contrastive segment loss. The first component  enables the algorithm to focus on challenging utterances. For  such utterances, the error detection component aims to identify the regions where mis-recognition occurs. Subsequently, by employing a contrastive segment loss, the algorithm aims to enhance recognition performance within these identified error regions.
 
 The algorithm works very similar to the EM algorithm \cite{bishop,Baum1967}  to minimise the errors seen w.r.t a particular utterance reference word sequence pair  $(\mathbf{x}_{1:T}, \mathbf{W}^r_{1:k})$.  To align with the EM algorithm's terminology, in the  ``E" step, the current parameter vector $\bm{\theta}^{\rm{old}}$ is used to find the posterior distribution $P_{\bm{\theta}^{\rm{old}}}( \mathbf{W}_{1:L}| \mathbf{x}_{1:T})$. The posterior distribution is then used to find the expected loss evaluated for some parameter vector $\bm{\theta}$. The expectation denoted by $\mathcal{Q}(\bm{\theta}^{\rm{old}}, \bm{\theta}) $ is given by:
\begin{align}
  \mathcal{Q}(\bm{\theta}^{\rm{old}}, \bm{\theta}) &= \sum_{\mathbf{W}_{1:L}} P_{\bm{\theta}^{\rm{old}}}( \mathbf{W}_{1:L}| \mathbf{x}_{1:T})  \ \ \mathcal{L}(  \bm{\theta},\mathbf{W}_{1:L}, \mathbf{W}^r_{1:k} |\mathbf{x}_{1:T}) \label{eqn:aux_fuc}
\end{align}
In the  ``M" step, a revised parameter estimate $\bm{\theta}^{\rm{new}}$ is determined by minimising the above function:
\begin{align*}
\bm{\theta}^{\rm{new}} = \argmin_{\bm{\theta}} \mathcal{Q}(\bm{\theta}^{\rm{old}}, \bm{\theta})
\end{align*}
The goal of the   ``M" step  is to reduce the loss associated with each hypothesis. The probability $P_{\bm{\theta}^{\rm{old}}}( \mathbf{W}_{1:L}| \mathbf{x}_{1:T})$ serves as a measure of how much emphasis the  approach should place on reducing errors associated with a specific $\mathbf{W}_{1:L}$. This enables the prioritisation of the most confusing hypotheses.  In the context of a deterministic tokenizer that maps  $\mathbf{W}_{1:L}$ to its associated word-piece sequence $\mathbf{l}_{1:U}$, $P_{\bm{\theta}^{\rm{old}}}( \mathbf{W}_{1:L}| \mathbf{x}_{1:T}) =  P_{\bm{\theta}^{\rm{old}}}( \mathbf{l}_{1:U}| \mathbf{x}_{1:T})$.  
This work utilises  $P_{\bm{\theta}^{\rm{old}}}( \mathbf{l}_{1:U}| \mathbf{x}_{1:T})$ to score each competing hypothesis and considers only the top N best alternatives due to the intractability of computing eqn(\ref{eqn:aux_fuc}) exactly. The following sections details the specific loss used in this paper.

\subsection{Error region detection \label{sec:segm}}
The proposed loss in this paper targets segments of the input sequence affected by ASR errors. The first step in designing this loss involves developing a procedure to identify regions in the audio where the ASR system performs poorly.   For a given input-reference word-piece label sequence pair $(\mathbf{x}_{1:T}, \mathbf{l}^{{r}}_{1:U})$, its CTC alignment $\hat{\bm{z}}^{r}_{1:T} (\bm{\theta})$ defines a unique segmentation of the encoder output sequence $\bm{e}_{1:T}$. This segmentation aligns the $i$-th token $\mathbf{l}^{r}_i$ with a unique sub sequence $\bm{e}_{t_i:t_{i+1}}$, such that $\mathbb{B}( \hat{\bm{z}}^{r}_{t_i:t_{i+1}}) = \mathbf{l}^{r}_i$. This procedure is exemplified by the following example. Consider the alignment $[ \emptyset,\emptyset,\mbox{`a'}, \emptyset, \emptyset, \mbox{`b'}] $ for an input sequence $\bm{x}_{1:6}$. Although CTC training marginalises over all possible alignments allowed by its latent discriminative model, it is well-known that CTC training converges to a dominant alignment \cite{bayctc,ctc_peaky}. In this example the observation that $\bm{z}_3 =$ `a' and $\bm{z}_6 =$ `b' implies that the CTC trained model emits the token `a' after processing $\bm{x}_{1:3}$ and token `b' after  processing $\bm{x}_{1:6}$.

The segmentation scheme proposed in this paper operates under the assumption that all the acoustic information required to predict `a' is captured by the encoder output sequence $\bm{e}_{1:3}$, where $\bm{e}_t = f(\bm{x}_{1:t}, \bm{\theta})$, and the information required to predict b' is contained within the encoder sequence $\bm{e}_{3:6}$ which operates on the frames $\bm{x}_{1:6}$. It is worth noting that the  scheme is dynamic (i.e., dependent on the choice of $\bm{\theta}$ and assumes the acoustic information from the right context frames is not required when predicting a token. Thus, this particular segmentation scheme will not be appropriate for encoders employing global attention to generate output at each time step i.e $\bm{e}_t = f(\bm{x}_{1:T}, \bm{\theta})$.

In the presence of a deterministic tokenizer, the above segmentation scheme can be extended to align each reference  word  $\mathbf{W}^{r}_k$ with a unique sub-sequence $\bm{e}_{t_k:t_{k+1}}$, thus allowing the identification of segments conforming to word boundaries. By comparing the Viterbi alignment of each competing hypothesis along the marked segments, the procedure  provides a framework to identify regions when the model's recognition deviates away from the intended words.

\subsection{Constrained CTC Graph}
To improve recognition of identified word segments, this work employs a contrastive loss  that utilises a constrained variant of  CTC's implicit discrete latent variable model  to represent $P({\mathbf{l}}_{1:u}| \mathbf{x}_{1:T})$, where $\mathbf{l}_{1:u}$ denotes word-pieces sequence associate with a word.   The original CTC loss formulation, designed for character-based models, interleaves label units with blanks in its discrete latent variable model (CTC graph) to accommodate character repetitions in generated label sequences. However, this design is unnecessary for word-piece models, as word-piece units typically do not repeat within a word. Therefore, this work during training adapts the CTC graph for word-piece models and models $P({\mathbf{l}}_{1:u}| \mathbf{x}_{1:T})$ with the following Markov random field \cite{bishop, MRK}:
\begin{align}
P({\mathbf{l}}_{1:u}| \mathbf{x}_{1:T}) &= \frac{1}{\mathcal{Z}} Q_{\bm{\theta}}({\mathbf{l}}_{1:u}| \mathbf{x}_{1:T}), \notag \\
&=  \frac{1}{\mathcal{Z}} \sum_{\bm{z}_{1:T}}    Q_{\bm{\theta}}(\bm{z}_{1:T}| \mathbf{x}_{1:T}) \cdot  {1}_{{\mathbf{l}}_{1:u}}(\bm{z}_{1:T} ).
\end{align}
In the context of this Markov random field, the function $ Q_{\bm{\theta}}(\bm{z}_{1:T}| \mathbf{x}_{1:T})$ factorises into a product of functions over maximal cliques in the CTC graph as follows:
\begin{align}
 Q_{\bm{\theta}}(\bm{z}_{1:T}| \mathbf{x}_{1:T}) &=   
   {1}(\bm{z}_{1:T} )   \prod_{t=1}^T  P_{\bm{\theta}}( \bm{z}_{t} |\mathbf{x}_{1:t}) \ q(\bm{z}_{t} | \bm{z}_{t-1}, T). \label{eqn:constctc}
\end{align}
The partition function $\mathcal{Z}$ presents a summation over all label sequences to ensure that the distribution is correctly normalised.  The introduction of the characteristic function $ {1}(\bm{z}_{1:T} )$ enforces an additional constraint on the CTC graph, prohibiting diagonal transition from a label token to the blank symbol unless its the last token in the label sequence. It results in constrained form of CTC graph where blanks are not allowed between word-pieces within a word. This results in the algorithm's time complexity being reduced to $\mathcal{O}(N^2T)$ from  $\mathcal{O}((2U+1)^2T)$, where $N <2U+1$. This work also enforces a time-invariant transition probability $q(\bm{z}_{t} | \bm{z}_{t-1}, T)$ to discourage consecutive blank transitions. For transitions involving non-blank tokens, equal probabilities are assigned to both horizontal and diagonal movements. However, when $\bm{z}_{t-1} =\emptyset$, the transition probability is $\frac{1}{T}$ for horizontal transition and $\frac{T-1}{T}$ for diagonal transition. 

\subsection{Contrastive Loss}
This work assumes the existence of a deterministic tokenizer $T$, which, for any given a pair $(\mathbf{W}_{1:L},\mathbf{l}_{1:U})$, facilitates the alignment of each word $\mathbf{W}_{k}$ with a unique sub-sequence $\mathbf{l}_{j_k:j_{k+1}}$. For a given $\mathbf{x}_{1:T}$, let $\hat{\mathbf{z}}^r_{1:T}$ denote the CTC alignment  associated with the reference word-piece sequence and  $\hat{\mathbf{z}}_{1:T}$ denote the CTC alignment associated with the word-piece sequence of an competing hypothesis. Additionally, let $\hat{\mathbf{z}}_{[t_k: t_{k+1}]}$ denote the segment within the CTC alignment that aligns with  $\mathbf{l}_{j_k:j_{k+1}}$ i.e $\mathbb{B}( \hat{\mathbf{z}}_{[t_k: t_{k+1}]}) = \mathbf{l}_{j_k:j_{k+1}}$. By defining 
$\mathbf{l}^e_{j_k:j_{k+1}}$ to  correspond to word-piece labels in the alignment $\hat{\mathbf{z}}_{[t_k: t_{k+1}]}$ that are not present within the alignment segment in $\mathbf{z}^r_{[t_k:t_{k+1}]}$,  the following segment-level loss function can now be delineated:

\begin{align}
&\mathcal{L}(  \bm{\theta},\mathbf{W}_{1:L}, \mathbf{W}^r_{1:k} |\mathbf{x}_{1:T}) =\mathcal{L}(  \bm{\theta},\mathbf{l}_{1:U}, \mathbf{l}^r_{1:U}|\mathbf{x}_{1:T}) \notag \\
&=\sum_{k=1}^K  {1}_{\mathbf{z}^r_{[t_k:t_{k+1}]}} (\mathbf{z}_{[t_k: t_{k+1}]} ) \log \dfrac { Q_{\bm{\theta}}(\mathbf{l}^e_{j_k:j_{k+1}}| \mathbf{x}_{[1: t_{k+1}]} ) }
{ Q_{\bm{\theta}}(\mathbf{l}^r_{[j_k: j_{k+1}]} |\mathbf{x}_{[1: t_{k+1}]}) } 
\end{align}

The characteristic function $ {1}_{\mathbf{z}^r_{[t_k:t_{k+1}]}}(\mathbf{z}_{[t_k: t_{k+1}]})$ equals to 0 if $ \mathbb{B}(\mathbf{z}_{[t_k: t_{k+1}]})=\mathbb{B}(\mathbf{z}^r_{[t_k: t_{k+1}]}) $ and  1 otherwise, and enables the identification of the specific audio segments where the model's recognition does not align with its predicted guess. 
The efficacy of the above segment level loss can be understood by studying its corresponding gradient update at time $t$. Let $\gamma^r_t$ represent the vector where the $j$-th entry signifies the sum of probabilities of all alignments in the constrained CTC graph of $\mathbf{l}^r_{[j_k: j_{k+1}]}$, assigning time $t$ to the $j$-th token and normalized by the sum of all allowed alignments. Similarly, let $\gamma_t$ denote the equivalent probability yielded at each time step from the constrained CTC graph of $\mathbf{l}^e_{j_k:j_{k+1}}$. Then,

\begin{equation}
   - \nabla   \mathcal{L}(  \bm{\theta})_t(j) = 
        \left\{ \begin{array}{ll}
             \gamma^r_t(j)   & \mbox{if} \ j \mbox{ in } \mathbf{l}^r_{[j_k: j_{k+1}]} ,\\
            \gamma^{ref}(j) - \gamma_t(j)  &\text{ if} \ j  \mbox { is } \emptyset,\\
           -  \ \gamma_t(j) & \text{if} \ j \mbox{ in }  \mathbf{l}^e_{j_k:j_{k+1}}, \\
           0 & \text{otherwise}.
        \end{array} \right.  \label{eqn:ft_grad}
    \end{equation}

The segment-level loss selectively updates certain nodes in the network at each time step: tokens aligned with the reference label sequence receive positive updates, boosting their probabilities, while tokens linked to insertion or substitution errors undergo negative updates to decrease their probabilities.


\section{Experiments \label{sec:exp}}

\subsection{Experimental Setup}
The algorithm's efficacy was assessed in improving streaming word-piece models trained with CTC for a fixed number of epochs on the 960-hour LibriSpeech dataset \cite{librispeech}, using the standard LibriSpeech test sets ('test-clean' and 'test-other'). Additionally, this work also evaluated the efficacy of the approach in further fine-tuning   a converged CTC+AED trained E2E model that had been first pre-trained on 600k hours of semi-supervised anonymised English dictation and assistant query data and then later fine-tuned  on 52K hours of supervised data from the same tasks.

To efficiently train the model within a fixed computational budget on both the LibriSpeech dataset and 600k hours dataset, this study focused on experimenting with a  conformer encoder paired with a bidirectional attention decoder consisting of 104 million parameters. The conformer encoder comprised of 12 blocks, each equipped with 8 attention heads. To optimise efficiency, the kernel size of the convolutional layers in the conformer was reduced to 8, and the acoustic feature sequence was sub-sampled by a factor of 6. The encoder was configured to utilise only the left context, enabling it to operate in a streaming fashion. The bidirectional transformer decoder consisted of 3 blocks, with each block featuring 8 attention heads in both the left and right decoders. The input features consisted of 80-channel Fbank features extracted from 25 ms windows shifted by 10 ms, with dither set to 0.1. Additionally, SpecAugment \cite{SpecAug} was employed to augment the training data for models trained on anonymised dictation and assistant tasks. The overall model configuration is comparable to the streaming model described in \cite{shams}.

On the LibriSpeech dataset, the impact of the segmentation strategy (Sec. \ref{sec:segm}) on the gains achieved through FDT was assessed. Two baseline models  were trained using CTC+AED loss; one using a streaming encoder and the other employing a  non-streaming one. Both baseline models were trained using a fixed computational budget: for 80 epochs using the Adam optimiser \cite{adam} on a single V100 GPU.  Discriminative training was performed for 1 epoch. The effectiveness of FDT was compared with MMI and MWER loss applied on top of the encoder, using N-best lists. Additionally, comparisons were made between the discriminative training approaches on further fine-tuning a streaming encoder that had been trained with CTC alone (with the decoder omitted), using the same training setup as the baseline. 

A second experiment was conducted to explore the advantages of additional fine-tuning with FDT on a streaming acoustic model trained using CTC+AED loss on a large-scale dataset from a different domain. The seed model was trained on a dataset comprised of dictation and assistant English requests in two stages. In the first stage, the model underwent pre-training with 600k hrs of semi-supervised data \cite{semi}. This pre-trained model  was then fine-tuned in the second stage using 52k hours of supervised data.  The fine-tuning was guided by the WER on the held out validation set.  Training was found to converge after 9 epochs. As a refined step, the chosen model then underwent additional fine-tuning on 2.5k hrs of randomly sampled data from the supervised set for 1 epoch. To make fair comparison, FDT was compared with MWER loss applied on top of the encoder during this additional fine-tuning phase.  The resultant models underwent evaluation on carefully curated dictation and assistant test sets, comprising 27 hours and 43 hours, respectively, with no overlap in speakers.

It should be mentioned that for all seed models whose training incorporated the AED loss, CTC+AED loss smoothing was applied during MWER and MMI training, whereas AED loss smoothing was used for FDT.

\subsection{Experimental Results \label{sec:expr}}

\begin{table}
          \centering
            \begin{tabular}{|c|c|c|l|l|} \hline 
         &      \multicolumn{2}{|c|}{\textbf{ls test clean}} &   \multicolumn{2}{|c|}{\textbf{ls test other}} \\ \hline 
         Loss&  ctc-prefix& joint decode& ctc-prefix&joint decode\\ \hline 
         CTC+AED&  4.1&  3.2& 10.22&8.52\\ \hline 
         +MMI&  4.03&  3.17& 10.12&8.37\\ \hline 
         +MWER&  4.04&  3.18& 10.22&8.44\\ \hline 
         +FDT&  \textbf{3.91}&  \textbf{3.12}& \textbf{10.04}&\textbf{8.34}\\ \hline
    \end{tabular}
         \caption{WER results with different discriminative approaches on fine-tuning a \textbf{streaming} CTC+AED trained model. }
          
    \label{tab:1}
      \end{table}

 \begin{table}
     \centering
     
     \begin{tabular}{|c|c|c|} \hline 
          Loss&  ls test clean& ls test other\\ \hline 
          CTC &  4.83& 11.84\\ \hline 
          + MMI&  4.83 & 11.84 \\ \hline 
          +MWER& 4.83 & 11.84 \\ \hline 
          +FDT&  \textbf{4.7}& \textbf{11.7}\\ \hline
     \end{tabular}
     
     \caption{ WER results with different discriminative approaches on fine-tuning a  CTC trained \textbf{streaming} encoder. }
     \label{tab:2}
 \end{table}

\begin{table}
    \centering
    \begin{tabular}{|c|c|c|l|l|} \hline 
         &  \multicolumn{2}{|c|}{\textbf{ls test clean}} &   \multicolumn{2}{|c|}{\textbf{ls test other}} \\ \hline 
         Loss&  ctc-prefix& j-decode& ctc-prefix&j-decode\\ \hline 
         CTC+AED&  3.16&  2.75& 7.81&6.66\\ \hline 
         +MMI&  3.16&  2.73& 7.8&6.66\\ \hline 
         +MWER&  \textbf{3.12}&  \textbf{2.7}& \textbf{7.7}&\textbf{6.56}\\ \hline 
         +FDT&  3.16&  2.75& 7.75&6.65\\ \hline
    \end{tabular}
   
         \caption{WER results with different discriminative approaches on fine-tuning a non-streaming CTC+AED trained model.}
    \label{tab:3}
\end{table}
 
Tables \ref{tab:1} and \ref{tab:3} compare the efficacy of FDT against MMI and MWER in further improving both streaming and non-streaming models trained with CTC+AED loss. These tables present WER comparisons with respect to CTC prefix beam search and the joint CTC-AED decoding framework \cite{ctcattention}, where CTC and AED scores are employed in the first pass.  Table \ref{tab:2} compares the discriminative training approaches in further improving a streaming CTC-trained encoder with respect to CTC prefix beam search. The model here in constrast to Table \ref{tab:1} was trained solely with CTC and no form CTC+AED loss smoothing was employed during discriminative training.

\begin{figure}
    \caption{ \label{fig:ctcspike} }
  \centering
      \includegraphics[scale =0.31]{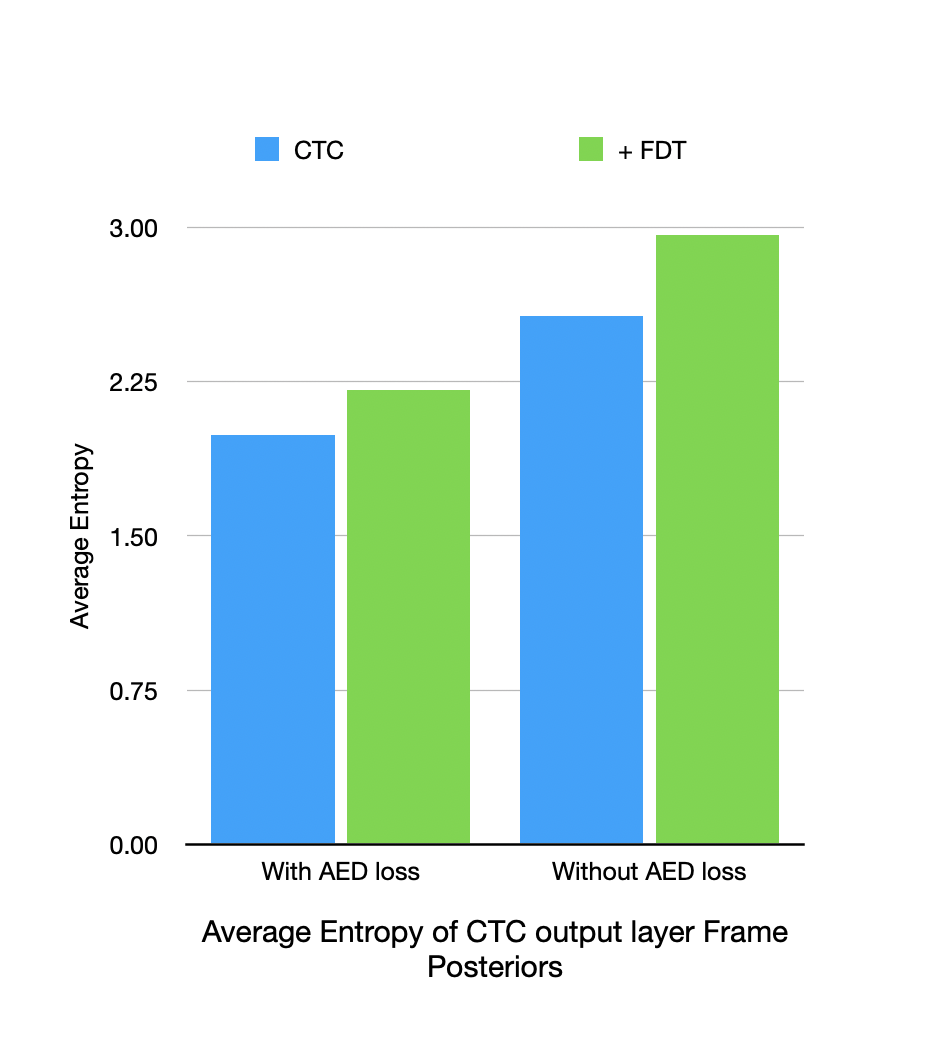}
  \end{figure}

From Tables \ref{tab:1} and \ref{tab:2}, FDT can be seen to the most effective in achieving the greatest WER reductions (WERR) from the streaming models. Comparison of WERs between the baseline models of Tables \ref{tab:1} and \ref{tab:2} revealed that model trained solely with CTC exhibited higher deletion errors on the test sets. Further analysis uncovered a correlation between the prevalence of higher deletions and the average entropy of the CTC softmax layer. Frame posteriors of CTC-only trained models exhibited lower average entropy compared to their CTC+AED counterparts, indicating greater `peaky behaviour' \cite{ctc_peaky}, i.e. a greater preference towards the dominant alignment. This  indicates that the AED loss plays a crucial role in reducing the `peaky behaviour' associated with CTC models. By imposing constraints (see Eqn (\ref{eqn:constctc})) on the blank transitions in the CTC search graph, FDT was observed to regulate the entropy of the trained models as shown in Figure \ref{fig:ctcspike}. The figure shows how aside from the AED loss, the FDT framework can also regulate the `peaky behaviour' associated with CTC models.  In addition, the investigation on the CTC-trained model revealed that CTC+AED loss smoothing was essential for achieving gains from MWER and MMI.

Table \ref{tab:3} shows how the strong assumptions made in segmentation strategy (Sec.\ref{sec:segm}) limits FDT's efficacy to only streaming models. On the non-streaming  LibriSpeech model, MWER was found to yield the greatest WERR.

\begin{table}
          \centering
          \begin{tabular}{|c|c|c|c|c|} \hline 
               Test set&  Decoding&  CTC&  + FDT&  + MWER\\ \hline 
            \multirow{2}{*}{Dictation}  &  ctc-prefix& 8.89 &\textbf{8.68} & 9.56 \\ 
              &  att. re-scoring&  7.84& \textbf{7.70} &   8.35 \\ \hline 
               \multirow{2}{*}{Assistant}&  ctc-prefix&  8.9 & \textbf{8.11} & 8.29 \\
               &  att. re-scoring& 6.91  &  \textbf{6.62}&  6.88 \\\hline
               
          \end{tabular}
       \caption{FDT vs MWER: Fine-tuning a \textbf{streaming} encoder trained on 600K hrs semi-supervised data, followed by 52K hrs of supervised data, using 2.5K hrs for fine-tuning.}
\label{tab:5}
\end{table}

When scaled to large-scale training with a different data distribution, the effectiveness of FDT in further improving the model is evident in Table \ref{tab:5}. This table presents comparisons between FDT and MWER with respect to CTC prefix decoding and attention re-scoring \cite{ctcattention} after additional fine-tuning using 2.5K hours of data for 1 epoch. FDT yields a relative 2\% improvement over the baseline model on the Dictation Task. On the Assistant Task, FDT achieves a 9\%  relative improvement in WER with respect to CTC prefix decoding (4\% with respect to attention re-scoring) over the baseline model, and a 2\%  relative improvement over additional WER fine-tuning.

\section{Conclusion \label{sec:conc}}
In summary,  this work introduces a novel framework for fine-tuning streaming models trained with either CTC or  CTC+AED loss. The proposed method is HMM-independent and lattice-free, with lower implementation overhead compared to current discriminative training approaches. This paper also presents an application of MWER on top of encoders.
In contrast to additional fine-tuning with MMI and MWER in an N-best list setting, FDT 
is shown to be comparatively more effective in achieving WERR  when fine-tuning is constrained to a fixed number of training steps or applied to a subset of the training data. This makes the method quite attractive in a setting where computation budget is fixed. This study also highlights the current limitations of the algorithm, particularly in scenarios where encoders use the entire acoustic context at each time step. Future work will focus on extending the efficacy of FDT to both streaming and non-streaming models.

\bibliographystyle{IEEEtran}

\end{document}